\documentclass[runningheads]{llncs}\usepackage[]{graphicx}\usepackage[]{color}
\makeatletter
\def\maxwidth{ %
  \ifdim\Gin@nat@width>\linewidth
    \linewidth
  \else
    \Gin@nat@width
  \fi
}
\makeatother

\definecolor{fgcolor}{rgb}{0.345, 0.345, 0.345}

\usepackage{framed}
\makeatletter
 {\par\unskip\endMakeFramed%
 \at@end@of@kframe}
\makeatother

\definecolor{shadecolor}{rgb}{.97, .97, .97}
\definecolor{messagecolor}{rgb}{0, 0, 0}
\definecolor{warningcolor}{rgb}{1, 0, 1}
\definecolor{errorcolor}{rgb}{1, 0, 0}
\newenvironment{knitrout}{}{} 

\usepackage{alltt}
\usepackage{amsmath,amssymb,array}
\usepackage{placeins}
\usepackage{xcolor}
\usepackage{graphicx}
\usepackage{xargs}
\usepackage[utf8]{inputenc}
\usepackage{booktabs}
\usepackage{todonotes}
\usepackage[square,numbers,sort&compress,sectionbib]{natbib}
\renewcommand\citet{\cite}

\usepackage[hyphens]{url}
\usepackage{hyperref}
\usepackage{colortbl}
\usepackage{csquotes} 
\usepackage{mathtools}
\usepackage{float} 
\usepackage{adjustbox}
\usepackage{dsfont}
\usepackage[ruled,vlined,linesnumbered]{algorithm2e}

\usepackage{dsfont}

\let\oldnl\nl
\newcommand{\nonl}{\renewcommand{\nl}{\let\nl\oldnl}}
\usepackage{cleveref}
\usepackage{autonum}
\renewcommand{\xi}[1][i]{x^{(#1)}}                                          
\ifdefined\N                                                                
\renewcommand{\N}{\mathds{N}}                                                
\else
  \newcommand{\N}{\mathds{N}}
\fi
\ifdefined\C
  \renewcommand{\C}{\mathds{C}}                                             
\else
  \newcommand{\C}{\mathds{C}}
\fi

\newcounter{mycomment}

\IfFileExists{upquote.sty}{\usepackage{upquote}}{}
\begin{document}

\title{Component-Wise Boosting of Targets for Multi-Output Prediction}
\titlerunning{}

\author{Quay~Au~\inst{1}, ~Daniel~Schalk~\inst{1}, ~Giuseppe~Casalicchio~\inst{1}, ~Ramona~Schoedel~\inst{2}, ~Clemens~Stachl~\inst{2}, ~Bernd~Bischl\inst{1}}
\authorrunning{Q. Au et al.}

\institute{Department of Statistics, Ludwig-Maximilians-University Munich,\\
  Ludwigstr. 33, 80539 Munich, Germany \\
  \email{\{quay.au, daniel.schalk, giuseppe.casalicchio, bernd.bischl\}@stat.uni-muenchen.de},\\
Department of Psychology, Methods and Assessment, Ludwig-Maximilians-University Munich,\\
Leopoldstr. 13, 80802 Munich, Germany\\
 \email{\{ramona.schoedel, clemens.stachl\}@psy.lmu.de}
}

\maketitle

\begin{abstract}
Multi-output prediction deals with the prediction of several targets of possibly diverse types.
One way to address this problem is the so called problem transformation method.
This method is often used in multi-label learning, but can also be used for multi-output prediction due to its generality and simplicity. In this paper, we introduce an algorithm that uses the problem transformation method for multi-output prediction, while simultaneously learning the dependencies between target variables in a sparse and interpretable manner. In a first step, predictions are obtained for each target individually. Target dependencies are then learned via a component-wise boosting approach. We compare our new method with similar approaches in a benchmark using multi-label, multivariate regression and mixed-type datasets.

\end{abstract}

\section{Introduction and Related Work}
\label{sec:introduction}
In traditional supervised learning, each instance is associated with one single outcome.
Multi-output (or multi-target) prediction is a supervised learning task, where multiple targets can be assigned to each observation.
In this learning problem, target variables can be of any kind (real-valued, discrete, categorical).

When all target variables are binary, this problem is known as multi-label classification \cite{Read2016, Shi2012, Tsoumakas2007, Zhang2014}.
Multi-label classification originated from text classification \cite{Schapire2000} and is increasingly being used in many different applications such as music categorization \cite{Li2012} or semantic scene categorization \cite{Boutell2003}.

On the other hand, if all target variables are real-valued, the multi-output prediction problem is known as multivariate regression. A broad overview of this topic can be found in \cite{Borchani2015}. Applications appear in many different fields, such as ecological modeling of multiple real-valued target variables describing the quality of vegetation, predicting wind noise (represented by several variables), or the estimation of multiple gas tank levels of a gas converter system.

Multi-output prediction can be seen as the most generalized and flexible form of learning to predict multiple targets, as it allows the target variables to be of mixed kind as well.
Important use cases for mixed target variables can be found in psychological research.  For instance, much work in the field of personality psychology is focused on the prediction of personality and demographic traits based on behavioral data \cite{Kosinski2013, Schoedel2018}. As traits like gender and age \cite{Chapman2007, Donnellan2008} have been found to be related to personality, it would be very useful to simultaneously predict personality via regression, gender via classification, and age via ordinal regression, instead of predicting them independently.

Currently, there are not many available methods that can handle learning tasks with objectives of different kinds (for an available method, see e.g. \cite{rfsrc2019}).
Instead of adapting existing methods to be able to handle more than one target, we will use the problem transformation method for predicting multiple targets instead.
For this, we will analyze the similarity-enforcing method \cite{Waegeman2018} of using predicted targets as feature representation, which has been studied in the multi-label community extensively \cite{Montanes2014, Read2011, Probst2017, Senge} and has been adapted to multivariate regression \cite{Borchani2015, SpyromitrosXioufis2012}.
We will define this method for the more general multi-output prediction problem and introduce a component-wise boosting approach for learning and visualizing the target dependencies.
Since the interpretability of black-box models has become an important topic in the machine learning community \cite{molnar2019}, we aimed for a method, that not only uses target dependencies for predictions, but also makes them easy to understand.

For general discussions about multi-output prediction in a broader context we refer to \cite{Waegeman2018, Xu2019ASO}.
Another method for multi-label classification, where label dependencies are learned in the form of rules, can be found here \cite{LozaMencia2016}.
The problem transformation method for multi-label learning is extensively discussed in many papers \cite{Zhang2014, Montanes2014, Probst2017, Read2011}.
This method has also been used for multivariate regression in \cite{Spyromitros-Xioufis2016, Borchani2015}.

Main contributions of this paper:
\begin{itemize}
  \item A formal definition of the problem transformation method for multi-output prediction problems.
  \item A novel method similar to the two-step stacking method, which allows interpretations and visualizations of target dependencies.
\end{itemize}

\section{Definition: Multi-Output Prediction}
\label{sec:definitions}

A multi-output prediction problem can be characterized by $n$ instances $\mathbf x^{(i)} \in \mathcal X$, $i = 1,...,n$ and $m$ targets $t_1, ..., t_m$. The relationship between an instance $\mathbf x^{(i)}$ and the target $t_j$ can be characterized by an one-dimensional score $y^{(i)}_j$, which can be nominal, ordinal, or real valued.
A multi-output prediction problem can thus be written as a dataset $\mathcal{D} = \{(\mathbf x^{(i)}, \mathbf y^{(i)})\}_{i=1,...,n}$, where the target variable is a vector $\mathbf y^{(i)} = \left(y^{(i)}_1, ..., y^{(i)}_m\right)$. This dataset can be portrayed in matrix form:

\begin{align}
D \hspace{5pt} \hat = \hspace{5pt}
\scalebox{.55}{
\begin{tabular}{|c|@{}|@{}|ccc|}
  \hline
Features & $t_1$ & ... & $t_m$ \\
  \hline
\cellcolor{black!10}  & \cellcolor{red!20} $y^{(1)}_1$ & \cellcolor{red!20} ... & \cellcolor{red!20} $y^{(1)}_m$ \\
  \cellcolor{black!10}  & \cellcolor{red!20} $y^{(2)}_1$ & \cellcolor{red!20} ... & \cellcolor{red!20} $y^{(2)}_m$ \\
  \cellcolor{black!10}  & \cellcolor{red!20} $y^{(3)}_1$ & \cellcolor{red!20} ... & \cellcolor{red!20} $y^{(3)}_m$ \\
  \cellcolor{black!10}  & \cellcolor{red!20} ... & \cellcolor{red!20} ... & \cellcolor{red!20} ... \\
  \cellcolor{black!10}  & \cellcolor{red!20} $y^{(n)}_1$ & \cellcolor{red!20} ... & \cellcolor{red!20} $y^{(n)}_m$ \\
   \hline
\end{tabular}
}
\end{align}

We can get the formal definition for multivariate regression by only allowing real values for $y^{(i)}_j$. By limiting $y^{(i)}_j$ to binary values 0 or 1, we get the formal definition for multi-label classification. However, since we do not need this limitation and want to deal with prediction problems with heterogeneous output spaces as well, we allow $y^{(i)}_j$ to be of any one-valued kind.
We call this problem \textbf{multi-output prediction}, which can be seen as a generalization of multi-label classification and multivariate regression.
We use the term multi-output prediction to refer to the general prediction task and only specify the terms multi-label classification, multivariate regression, or mixed-type prediction, if we specifically relate to them.

%

%
%



\section{Measuring Performance in Multi-Output Prediction Problems}
\label{sec:perfmeasure}

For traditional single-target machine learning problems, performance measurement is intuitive and there are many metrics like accuracy, F-measure, AUC for classification, or the mean squared error (MSE), mean absolute error (MAE) for regression.
Once we have multiple target variables, measuring performance becomes non-trivial.

There are many ways of handling this problem.
First, we can compare the actual target vector $\mathbf y = \left(y_1, ..., y_m\right)$ with the predicted target vector $\hat{\mathbf y} = \left(\hat y_1, ..., \hat y_m\right)$ and then calculate one performance metric. Many performance measures have been constructed this way for multi-label classification and multivariate regression problems \cite{Zhang2014, Borchani2015}.

For multi-label learning, an example would be the so called Hamming-loss, which compares the predicted labels with the actual labels:
\begin{eqnarray}
\text{HL}(\mathbf y, \hat{\mathbf y}) =  \frac{1}{m}\sum_{j = 1}^m \mathds{1}_{\{y_j \neq \hat y_j\}}
\end{eqnarray}
This value is calculated instance-wise and the performance of a test set is the mean Hamming-loss of each instance.

There are many more multi-label performance measures like $\text{Subset}_{01}$-loss, Accuracy, Precision, or Ranking-loss (see \cite{Zhang2014}), which can be defined intuitively, because of the binary structure of multi-label learning problems.

For multivariate regression, an example is the multivariate mean squared error MMSE, which is the mean MSE of every target:
\begin{eqnarray}
\text{MMSE}(\mathbf y, \hat{\mathbf y}) = \frac{1}{m}\sum_{j = 1}^m (y_j - \hat y_j)^2
\end{eqnarray}

Having only regression tasks for every target, many multivariate performance metrics can be defined (see e.g. \cite{Borchani2015}).
When using such metrics for multivariate regression problems, one should pay attention to the value range of the target variable.
Targets with larger value ranges have more influence on the metric than targets with smaller value ranges.
One possible way of handling this problem is to standardize the target values.

However, in the more generalized multi-output prediction problem, calculating one single performance value out of possible mixed target spaces is not trivial.
Note, that many multi-label and multivariate regression performance metrics are a weighted sum of performance metrics for each target.
We could write a general performance metric $\mathcal L$ like this:
\begin{eqnarray}
\mathcal L(\mathbf y, \hat{\mathbf y}) = \sum_{j = 1}^m \lambda_j \mathcal L_j(y_j, \hat y_j)
\end{eqnarray}
The Hamming-loss and the MMSE are just special cases of this more general performance metric.

Since datasets with mixed target spaces can differ very much and classification performance metrics are combined with regression performance metrics during evaluation, a general definition of a performance metric is infeasible and should thus be left to the user.
One could also handle multi-output prediction problems as multi-objective optimization problems, where trade-offs between multiple (possibly conflicting) objectives (such as minimizing the MSE for a regression target and maximizing the AUC for a classification target) need to be considered. For multi-label classification this was discussed in \cite{Shi2012}.

Nevertheless, a further motivation to consider multi-output prediction methods instead of modeling each target independently is that improvements can be made for each target respectively.
Each target can be treated independently and we can analyse whether more complex methods are feasible for each target.


For problems with mixed target variables, we will focus on target wise comparisons and use the mean classification error for classification problems and  the mean squared error for regression problems. For multi-label classification and multivariate regression we will also report the Hamming-loss and MMSE.

\section{Learning Target Dependencies}

There are two main ways to model problems with more than one target, that are extensively studied in the multi-label community \cite{Zhang2014, Probst2017} and could also be applied to  multi-output prediction. One of them is the \textit{algorithm adaptation method}, which aims at adapting existing algorithms to handle multiple outputs \cite{Zhang2007}. The other one is called \textit{problem transformation method} and aims to transform the multi-label learning problem into more established one-target prediction problems \cite{Zhang2014, Read2011, Montanes2014}. The problem transformation method has the advantage that any already established one-target machine learning model can be used.

In this paper, we will focus on the problem transformation method and how to use it for multi-output prediction problems.
Originally used in the multi-label community, these methods were adapted to multivariate regression in \cite{SpyromitrosXioufis2012}.
The idea of modeling target dependencies by using other target information as features is not restricted by the type of outputs and can thus be used for multi-output prediction problems as well.

\subsection{Independent Models (IM)}

The easiest problem transformation method (called binary relevance method in the multi-label community) is to use one model for each target independently and to combine the predictions afterwards. Target dependencies are thus not being considered when using independent models.

Given a dataset $\mathcal{D} = \{(\mathbf x^{(i)}, \mathbf y^{(i)})\}_{i=1,...,n}$, with target $\mathbf y^{(i)} = \left(y^{(i)}_{1}, ..., y^{(i)}_{m}\right)$ and $m$ (possibly mixed) targets, we train $m$ models $f_{j}$ for each target independently:
\begin{eqnarray}
\label{independentmodels}
\text{For $j = 1,...,m$ train model $f_j$ on }
\scalebox{.55}{
\begin{tabular}{|c|@{}|@{}|c|}
  \hline
Features & $t_j$ \\
  \hline
\cellcolor{black!10}  & \cellcolor{red!20} $y^{(1)}_{j}$ \\
  \cellcolor{black!10}  & \cellcolor{red!20} $y^{(2)}_{j}$ \\
  \cellcolor{black!10}  & \cellcolor{red!20} $y^{(3)}_{j}$ \\
  \cellcolor{black!10}  & \cellcolor{red!20} ... \\
  \cellcolor{black!10}  & \cellcolor{red!20} $y^{(n)}_{j}$ \\
   \hline
\end{tabular}
}
\end{eqnarray}
A new observation $x_{\text{new}}$ will get the prediction $f(x_{\text{new}}) = \left(f_1(x_{\text{new}}), ..., f_m(x_{\text{new}})\right)$.

\subsection{Stacking (STA): Using Targets as Features}
\label{sec:stacking}
One way to model target variable dependencies is to use target variables as features.
A distinction can be made between different ways in which these target variables are being modeled.
For instance, the real target values can be used as features, since these are available during training time.
Examples would be the classifier chains \cite{Read2011} or dependent binary relevance \cite{Montanes2014}.
The alternative would be to create predicted target values by using an inner cross-validation loop (e.g. nested stacking \cite{Senge}, stacking \cite{Waegeman2018, Montanes2014}).
A comparison between these methods is discussed in \cite{Probst2017}.
In this paper, however, we will discuss the stacking method in more detail.
After fitting the same independent models (\ref{independentmodels}), as they are needed at prediction time, we obtain predicted targets $\{\hat{y}^{(i)}_{j}\}_{i = 1,...,n}$ through an inner cross-validation strategy:

\begin{eqnarray}
\label{cvpred}
\text{For $j = 1,...,m$ use inner-CV on }
\scalebox{.55}{
\begin{tabular}{|c|@{}|@{}|c|}
  \hline
Features & $t_j$ \\
  \hline
\cellcolor{black!10}  & \cellcolor{red!20} $y^{(1)}_{j}$ \\
  \cellcolor{black!10}  & \cellcolor{red!20} $y^{(2)}_{j}$ \\
  \cellcolor{black!10}  & \cellcolor{red!20} $y^{(3)}_{j}$ \\
  \cellcolor{black!10}  & \cellcolor{red!20} ... \\
  \cellcolor{black!10}  & \cellcolor{red!20} $y^{(n)}_{j}$ \\
   \hline
\end{tabular}
} \text{ to obtain } \scalebox{.55}{
\begin{tabular}{|c|}
  \hline
$\hat{t}_j$ \\
  \hline
\cellcolor{black!20} $\hat{y}^{(1)}_{j}$ \\
  \cellcolor{black!20} $\hat{y}^{(2)}_{j}$ \\
  \cellcolor{black!20} $\hat{y}^{(3)}_{j}$ \\
  \cellcolor{black!20} ... \\
  \cellcolor{black!20} $\hat{y}^{(n)}_{j}$ \\
   \hline
\end{tabular}
}
\end{eqnarray}
The inner cross-validation strategy can become resource-intensive, as many models have to be fit.
Hence, a trade-off between a sufficient cross-validation strategy and available computing resources needs to be made.

In a next step, these predicted target variables are used to extend the feature space, and a second set of models is fit for each target:

\begin{eqnarray}
\label{stacking}
\text{For $j = 1,...,m$ train model $g^{\text{sta}}_j$ on}
\scalebox{.55}{
\begin{tabular}{|cccc|@{}|@{}|c|}
  \hline
Features & $\hat t_1$ & ... & $\hat t_m$ &  $t_j$\\
  \hline
\cellcolor{black!10}  & \cellcolor{black!20} $\hat y^{(1)}_{1}$ & \cellcolor{black!20} ... & \cellcolor{black!20}  $\hat y^{(1)}_{m}$ &\cellcolor{red!20} $y^{(1)}_{j}$\\
  \cellcolor{black!10}  & \cellcolor{black!20} $\hat y^{(2)}_{1}$ & \cellcolor{black!20} ... & \cellcolor{black!20}  $\hat y^{(2)}_{m}$ &\cellcolor{red!20} $y^{(2)}_{j}$\\
  \cellcolor{black!10}  & \cellcolor{black!20} $\hat y^{(3)}_{1}$ & \cellcolor{black!20} ... & \cellcolor{black!20}  $\hat y^{(3)}_{m}$ &\cellcolor{red!20} $y^{(3)}_{j}$\\
  \cellcolor{black!10}  & \cellcolor{black!20} ... & \cellcolor{black!20} ... & \cellcolor{black!20} ...& \cellcolor{red!20} ...\\
  \cellcolor{black!10}  & \cellcolor{black!20} $\hat y^{(n)}_{1}$ & \cellcolor{black!20} ... & \cellcolor{black!20}  $\hat y^{(n)}_{m}$& \cellcolor{red!20} $y^{(m)}_{j}$ \\
   \hline
\end{tabular}
}
\end{eqnarray}

At prediction time, we first get predicted targets with independent models, which are then added to the new observation: $x^*_\text{new} = x_\text{new} \cup f(x_\text{new})$. The final prediction is $g^{\text{sta}}(x^*_\text{new}) = (g^{\text{sta}}_1(x^*_\text{new}), ..., g^{\text{sta}}_m(x^*_\text{new}))$.



%



\subsection{Component-wise Multi-output Boosting (CMOB)}
For our novel method, we propose to use component-wise boosting to learn the target dependency structure. As for most machine learning models, the aim of component-wise boosting is to minimize the empirical risk:
$$
\mathcal{R}_\text{emp}(g_j) = \frac{1}{n} \sum\limits_{i = 1}^n L(y, g_j)
$$

%
%
%

Component-wise boosting, also called model-based boosting, generalizes the boosting framework to multiple base-learners \cite{buhlmann2003boosting}. For each boosting iteration $k = 1, \dots, m_\text{stop}$ the algorithm selects one base-learner $b^{[k]}_l$ out of a space of base-learners $B = \{b_1, \dots, b_\tau\}$ by fitting them all to the pseudo residuals and choosing the one with the smallest sum of squared errors. This improves the empirical risk $\mathcal{R}_\text{emp}(g_j^{[k]})$ of the current model $g_j^{[k]}$ which is computed via stage-wise additive modeling with a learning rate $\nu \in (0,1)$:
\begin{align}
g_j^{[k]}(x) &= g_j^{[0]}(x)   + \nu \sum\limits_{i = 1}^k b^{[i]}_l(x) \\
             &= g_j^{[k-1]}(x) + \nu b^{[k]}_l(x)
\end{align}

For our purpose, numerical features are included as linear effect $b_l^{[k]}(x) = \theta_{0l}^{[k]} + \theta_{1l}^{[k]}\hat{t}_{\text{sel}(k)}$, where $\text{sel}(k)$ is a mapping from iteration $k$ to the selected feature. For categorical features each group is added as single one-hot coded base-learner $b_l^{[k]}(x) = \theta_{0l}^{[k]}$ that just includes an intercept for that group. Boosting these kind of base-learners maintains interpretability because of the additive structure of the model and the repeated selection of equal base-learners.

An important property of component-wise boosting is the intrinsic feature selection. This is achieved by selecting just one base-learner per iteration. After training $m_\text{stop}$ iterations we get a subset of all features that are required to predict the target. This provides information about the importance of each feature.
In our multi-output prediction case we use this internal feature selection to learn which predicted target variables are required to explain the target $t_j$.

To go one step further we would also like to know which of the selected features are more important than others. Therefore, we can again use the additive structure of component-wise boosting to calculate a feature importance for all selected features. After boosting $m_\text{stop}$ iterations we calculate the feature importance as the sum of the empirical risk improvements achieved by selecting the $p$-th feature:

$$
\text{vip}_p = \sum\limits_{k = 1}^{m_\text{stop}} \left(\mathcal{R}_\text{emp}\left(g_j^{[m_\text{stop}-1]}\right) - \mathcal{R}_\text{emp}\left(g_j^{[m_\text{stop}]}\right)\right)\mathds{1}_{\{p = \text{sel}(k)\}}
$$

One requirement for calculating meaningful feature importance scores is to choose an adequate $m_\text{stop}$ which can be done, by using early stopping. This stops the procedure if the relative improvement
$$\frac{\mathcal{R}_\text{emp}\left(g_j^{[k-1]}\right) - \mathcal{R}_\text{emp}\left(g_j^{[k]}\right)}{\mathcal{R}_\text{emp}\left(g_j^{[k-1]}\right)}$$
of the empirical risk consecutively falls below a pre-defined value $\epsilon$.
We chose component-wise boosting over other methods, which produce sparse and interpretable models (like ridge regression), because of the flexibility of the choice of the base-learners. Non-linear effects can easily be modeled using splines as base learners \cite{schmid2008boosting}.

We now introduce \textit{Component-Wise Multi-output Boosting} (CMOB) (see algorithm \ref{algo}). The idea is to use component-wise boosting to learn the target dependencies in a sparse and interpretable manner.
CMOB aims at modeling target dependencies through a dataset of predicted target variables $\hat{Y}$, just like the stacking algorithm (see section \ref{sec:stacking}).

One difference is that in our algorithm the original features are omitted, because we are only interested in the interactions between the target variables. Interactions between predicted target variables and features are thus not modeled.

Given the dataset $\hat{Y}$ of predicted target variables (obtained by (\ref{cvpred})), we train component-wise boosting models for each target:

\begin{eqnarray}
\text{For j = 1,...,m train component-wise boosting model $g_j$ on }
\scalebox{.55}{
\begin{tabular}{|ccc|@{}|@{}|c|}
  \hline
$\hat{t}_1$ & ... & $\hat{t}_m$ & $t_j$ \\  \hline
\cellcolor{black!20} $\hat{y}^{(1)}_{1}$ & \cellcolor{black!20} ... & \cellcolor{black!20} $\hat{y}^{(1)}_{m}$ & \cellcolor{red!20} ${y}^{(1)}_{j}$ \\
  \cellcolor{black!20} $\hat{y}^{(2)}_{1}$ & \cellcolor{black!20} ... & \cellcolor{black!20} $\hat{y}^{(2)}_{m}$ & \cellcolor{red!20} ${y}^{(2)}_{j}$ \\
  \cellcolor{black!20} $\hat{y}^{(3)}_{1}$ & \cellcolor{black!20} ... & \cellcolor{black!20} $\hat{y}^{(3)}_{m}$ & \cellcolor{red!20} ${y}^{(3)}_{j}$ \\
  \cellcolor{black!20} ... & \cellcolor{black!20} ... & \cellcolor{black!20} ... & \cellcolor{red!20} ... \\
  \cellcolor{black!20} $\hat{y}^{(n)}_{1}$ & \cellcolor{black!20} ... & \cellcolor{black!20} $\hat{y}^{(n)}_{m}$ & \cellcolor{red!20} ${y}^{(n)}_{j}$ \\
   \hline
\end{tabular}
}
\end{eqnarray}

A new observation $x_{\text{new}}$ will be predicted in a two-step procedure:
\begin{itemize}
  \item[1)] Use independent models (\ref{independentmodels}) to create predicted targets:
  $$f(x_{\text{new}}) = \left(f_1(x_{\text{new}}), ..., f_m(x_{\text{new}})\right)$$
  \item[2)] Use boosting models to create final predictions:
  $$g\left(f(x_{\text{new}})\right) = \left(g_1(f(x_{\text{new}})), ..., g_m(f(x_{\text{new}}))\right)$$
\end{itemize}

\begin{algorithm}[H]
  \label{algo}
  \SetKwData{Left}{left}\SetKwData{This}{this}\SetKwData{Up}{up}
  \SetKwFunction{Union}{Union}\SetKwFunction{FindCompress}{FindCompress}
  \SetKwInOut{Input}{input}\SetKwInOut{Output}{output}
  \Input{Dataset $\mathcal{D} = \{(\mathbf x^{(i)}, \mathbf y^{(i)})\}_{i=1,...,n}$ with targets $\mathbf y^{(i)} = \left(y^{(i)}_{1}, ..., y^{(i)}_{m}\right)$}
  \Output{Prediction model for CMOB.}
  \tcp{First level models and predictions:}
  \For{$j \in \{1, ..., m\}$}{
   subset data: $D_j = \{(\mathbf x^{(i)}, y^{(i)}_{j})\}_{i=1,...,n}$\;
   train model $f_j$ on $D_j$\;
   get cross-validated predictions $\{\hat{y}^{(i)}_{j}\}_{i = 1,...,n}$\;
  }
  \tcp{Create second level data:}
  Define $\hat{Y}:=\{\hat{y}^{(i)}_{j}\}_{i = 1,...,n, j = 1,...,m}$\;
  \tcp{Run component-wise boosting:}
  \For{$j \in \{1, ..., m\}$}{
    train component-wise boosting model $g_j$: $y_j \sim \hat{Y}$\;
  }
  \caption{Component-Wise Multi-output Boosting (CMOB)}
\end{algorithm}


\section{Benchmark}
\label{sec:benchmark}

\subsection{Datasets}
\label{sec:datasets}
We use openly available datasets, that can be downloaded from OpenML \cite{OpenML2013, OpenMLR2017}.
Since datasets for mixed-type prediction are quite uncommon, we mainly used multi-label and multivariate regression datasets.
We have limited the number of targets to a maximum of 7 in order to keep the computing time reasonable and the visualizations more understandable.
The multi-label classification and multivariate regression datasets are described in detail in \cite{mulan, Spyromitros-Xioufis2016, Probst2017}.
The mixed-type datasets are both personality prediction datasets \cite{biel2013youtube, Schoedel2018}.
See table \ref{tab:datasets} for more details on the datasets.

\begin{table}[ht]
\centering
\begin{tabular}{rrrrrrr}
  \toprule
Dataset & n & nfeats & ntargets & Type & Reference & Data ID \\ 
  \midrule
emotions & 593 &  72 &   6 & multilabel & \href{http://ismir2008.ismir.net/papers/ISMIR2008_275.pdf}{link} & 41545 \\ 
  image & 2000 & 135 &   5 & multilabel & \href{http://citeseerx.ist.psu.edu/viewdoc/download?doi=10.1.1.538.9597&rep=rep1&type=pdf}{link} & 41546 \\ 
  reuters & 2000 & 243 &   7 & multilabel & \href{http://cse.seu.edu.cn/people/zhangml/files/icdm08.pdf}{link} & 41547 \\ 
  scene & 2407 & 294 &   6 & multilabel & \href{http://citeseerx.ist.psu.edu/viewdoc/download?doi=10.1.1.113.3921&rep=rep1&type=pdf}{link} & 41548 \\ 
  andro &  49 &  30 &   6 & multiv. regr. & \href{https://link.springer.com/article/10.1007%2Fs10994-016-5546-z}{link} & 41549 \\ 
  atp1d & 337 & 411 &   6 & multiv. regr. & \href{https://link.springer.com/article/10.1007%2Fs10994-016-5546-z}{link} & 41550 \\ 
  atp7d & 296 & 411 &   6 & multiv. regr. & \href{https://link.springer.com/article/10.1007%2Fs10994-016-5546-z}{link} & 41551 \\ 
  edm & 154 &  16 &   2 & multiv. regr. & \href{https://link.springer.com/content/pdf/10.1023/A:1007365207130.pdf}{link} & 41552 \\ 
  enb & 768 &   8 &   2 & multiv. regr. & \href{https://link.springer.com/article/10.1007%2Fs10994-016-5546-z}{link} & 41553 \\ 
  jura & 359 &  15 &   3 & multiv. regr. & \href{https://link.springer.com/article/10.1007%2Fs10994-016-5546-z}{link} & 41554 \\ 
  scpf & 1137 &  23 &   3 & multiv. regr. & \href{https://link.springer.com/article/10.1007%2Fs10994-016-5546-z}{link} & 41555 \\ 
  sf1 & 323 &  10 &   3 & multiv. regr. & \href{http://archive.ics.uci.edu/ml/datasets/Solar+Flare}{link} & 41556 \\ 
  sf2 & 1066 &  10 &   3 & multiv. regr. & \href{http://archive.ics.uci.edu/ml/datasets/Solar+Flare}{link} & 41557 \\ 
  slump & 103 &   7 &   3 & multiv. regr. & \href{https://link.springer.com/article/10.1007%2Fs10994-016-5546-z}{link} & 41558 \\ 
  youtube & 404 &  25 &   6 & mixed-type & \href{https://ieeexplore.ieee.org/document/6331531}{link} & 41559 \\ 
  sens & 257 & 222 &   4 & mixed-type & \href{https://osf.io/v4xrf/}{link} & 41560 \\ 
   \bottomrule
\end{tabular}
\caption{Benchmark datasets which are available on OpenML and can be found via the data ID. The link is a hyperlink to the original source or paper.} 
\label{tab:datasets}
\end{table}

\subsection{Benchmark}
\label{sec:datasets}
To analyze the potential of learning the target dependency structure, we compare the performance of the proposed CMOB algorithm with a stacking model (STA), which uses all other predicted labels as features. We compare these algorithms with independent models (IM) as baseline. See table \ref{tab:benchmarksettings} for an overview of the benchmark settings.

For CMOB we use linear base learners for the underlying component-wise boosting algorithm with a maximum number of 10000 iterations. Since we strive for sparse models, we have applied an early-stopping strategy. The boosting process stops when no improvement of at least $0.01\%$ has been achieved for 5 consecutive iterations.

As one-target algorithms for classification and regression we will use random forests \cite{breiman2001random}, as they typically perform well in many different scenarios without the need of tuning hyperparameters.

Performance will be evaluated with an outer 10-fold-cross-validation strategy. For classification tasks we will use the mean misclassification error (mmce) as performance metric. For regression tasks we will use the mean squared error (MSE) of the standardized target values (test set target values are standardized using mean and standard deviation of the respective training sets).
In the inner training sets, the predicted targets are created with an inner 10-fold-cross-validation strategy.
The outer test sets are only used for prediction and performance evaluation.
And finally, the models are trained on the whole datasets. For full reproducibility, the benchmark code is available here \cite{Au2019}.

\begin{table}[ht]
\centering
\begin{tabular}{lr}
  \toprule
Multi-output algorithms & IM, STA, CMOB \\
  Outer resampling strategy & 10 fold cv \\
  Resampling strategy for creating predictions & 10 fold cv \\
  One target regression learner & Random Forest \\
  One target classification learner & Random Forest \\
  Classification measure & mmce \\
  Regression measure & mse (of normalized target values) \\
  Base-learners & Linear \\
  Maximum iterations for boosting & 10000 \\
  Early stopping strategy & No 0.01\% improvement for 5 iterations. \\
   \bottomrule
\end{tabular}
\caption{Benchmark settings for independent models (IM), stacking (STA), and component-wise multi-output boosting (CMOB) with linear base learners.
We only added linear base learners, because splines are harder to interpret and some internal experiments with splines have indicated that these tend to overfit and have a worse generalization error.}
\label{tab:benchmarksettings}
\end{table}

\subsection*{Results}
We summarized the results of the benchmark in table \ref{tab.mtr}.
Reported values are mean values (over the outer test sets) of MSE or mmce (depending on the task) for each dataset and each target.
For multi-label classification tasks we also included the Hamming-loss (HL) and for multivariate regression tasks the multivariate mean squared error (MMSE).

CMOB could not improve the overall Hamming-loss of the multi-label datasets used in this benchmark. Looking at the performance values for each target individually, we can see that for some targets (e.g. $t_1$ for the dataset \textit{emotions}) CMOB could improve the mmce, but for others (e.g. $t_3$ for the dataset \textit{image}) our algorithm did worse. However, the stacking algorithm (STA) neither performed very well on the multi-label datasets and could only improve the Hamming-loss on the \textit{image} dataset by a small margin. Independently modeling each target individually seems to be a strong baseline here.

More interesting is the performance of CMOB for the multivariate regression datasets. For 4 (\textit{andro, jura, sf1, slump}) of the 10 multivariate regression datasets, CMOB could improve the MMSE over independent models. In 2 of these tasks, CMOB could even beat the stacking algorithm. Looking at each target variables individually, we can see that CMOB performs comparably well to the stacking algorithm, showing improvements over the MSE, when stacking also improves over independent models (with some exceptions e.g. $t_3$ from the dataset \textit{sf2}).

For the mixed-type dataset \textit{youtube} and \textit{sens} we can see improvements for some targets when using CMOB. This suggests that the use of multi-output methods can be useful for personality prediction.

Based on datasets for which considerable improvements have been achieved, we show the interpretations of the target dependencies in the following section.




\begin{table}[ht]
\centering
\begin{tabular}{llrrrrrrrrr}
  \toprule
Dataset & Algorithm & $t_1$ & $t_2$ & $t_3$ & $t_4$ & $t_5$ & $t_6$ & $t_7$ & MMSE & HL \\ 
  \midrule
\cellcolor{gray!25} andro & \cellcolor{gray!25} IM & \cellcolor{gray!25} 0.12 & \cellcolor{gray!25} 0.26 & \cellcolor{gray!25} 0.18 & \cellcolor{gray!25} 0.18 & \cellcolor{gray!25} 0.39 & \cellcolor{gray!25} 0.4 & \cellcolor{gray!25}  & \cellcolor{gray!25} 0.26 & \cellcolor{gray!25}  \\ 
  \cellcolor{gray!25} andro & \cellcolor{gray!25} STA & \cellcolor{gray!25} $\mathbf{0.08}$ & \cellcolor{gray!25} $\mathbf{0.21}$ & \cellcolor{gray!25} $\mathbf{0.11}$ & \cellcolor{gray!25} $\mathbf{0.11}$ & \cellcolor{gray!25} $\mathbf{0.32}$ & \cellcolor{gray!25} $\mathbf{0.33}$ & \cellcolor{gray!25}  & \cellcolor{gray!25} $\mathbf{0.19}$ & \cellcolor{gray!25}  \\ 
  \cellcolor{gray!25} andro & \cellcolor{gray!25} CMOB & \cellcolor{gray!25} $\mathbf{0.08}$ & \cellcolor{gray!25} $\mathbf{0.21}$ & \cellcolor{gray!25} $\mathbf{0.11}$ & \cellcolor{gray!25} $\mathbf{0.11}$ & \cellcolor{gray!25} 0.35 & \cellcolor{gray!25} 0.36 & \cellcolor{gray!25}  & \cellcolor{gray!25} 0.2 & \cellcolor{gray!25}  \\ 
  \cellcolor{gray!5} atp1d & \cellcolor{gray!5} IM & \cellcolor{gray!5} $\mathbf{0.22}$ & \cellcolor{gray!5} $\mathbf{0.17}$ & \cellcolor{gray!5} $\mathbf{0.2}$ & \cellcolor{gray!5} $\mathbf{0.09}$ & \cellcolor{gray!5} $\mathbf{0.21}$ & \cellcolor{gray!5} $\mathbf{0.08}$ & \cellcolor{gray!5}  & \cellcolor{gray!5} $\mathbf{0.16}$ & \cellcolor{gray!5}  \\ 
  \cellcolor{gray!5} atp1d & \cellcolor{gray!5} STA & \cellcolor{gray!5} $\mathbf{0.22}$ & \cellcolor{gray!5} 0.18 & \cellcolor{gray!5} $\mathbf{0.2}$ & \cellcolor{gray!5} $\mathbf{0.09}$ & \cellcolor{gray!5} $\mathbf{0.21}$ & \cellcolor{gray!5} $\mathbf{0.08}$ & \cellcolor{gray!5}  & \cellcolor{gray!5} $\mathbf{0.16}$ & \cellcolor{gray!5}  \\ 
  \cellcolor{gray!5} atp1d & \cellcolor{gray!5} CMOB & \cellcolor{gray!5} $\mathbf{0.22}$ & \cellcolor{gray!5} 0.19 & \cellcolor{gray!5} $\mathbf{0.2}$ & \cellcolor{gray!5} $\mathbf{0.09}$ & \cellcolor{gray!5} $\mathbf{0.21}$ & \cellcolor{gray!5} $\mathbf{0.08}$ & \cellcolor{gray!5}  & \cellcolor{gray!5} $\mathbf{0.16}$ & \cellcolor{gray!5}  \\ 
  \cellcolor{gray!25} atp7d & \cellcolor{gray!25} IM & \cellcolor{gray!25} 0.33 & \cellcolor{gray!25} $\mathbf{0.29}$ & \cellcolor{gray!25} 0.28 & \cellcolor{gray!25} 0.14 & \cellcolor{gray!25} 0.36 & \cellcolor{gray!25} $\mathbf{0.12}$ & \cellcolor{gray!25}  & \cellcolor{gray!25} $\mathbf{0.25}$ & \cellcolor{gray!25}  \\ 
  \cellcolor{gray!25} atp7d & \cellcolor{gray!25} STA & \cellcolor{gray!25} $\mathbf{0.32}$ & \cellcolor{gray!25} 0.3 & \cellcolor{gray!25} $\mathbf{0.27}$ & \cellcolor{gray!25} $\mathbf{0.13}$ & \cellcolor{gray!25} $\mathbf{0.35}$ & \cellcolor{gray!25} $\mathbf{0.12}$ & \cellcolor{gray!25}  & \cellcolor{gray!25} $\mathbf{0.25}$ & \cellcolor{gray!25}  \\ 
  \cellcolor{gray!25} atp7d & \cellcolor{gray!25} CMOB & \cellcolor{gray!25} 0.33 & \cellcolor{gray!25} 0.34 & \cellcolor{gray!25} 0.28 & \cellcolor{gray!25} $\mathbf{0.13}$ & \cellcolor{gray!25} 0.37 & \cellcolor{gray!25} $\mathbf{0.12}$ & \cellcolor{gray!25}  & \cellcolor{gray!25} 0.26 & \cellcolor{gray!25}  \\ 
  \cellcolor{gray!5} edm & \cellcolor{gray!5} IM & \cellcolor{gray!5} 0.41 & \cellcolor{gray!5} 0.45 & \cellcolor{gray!5}  & \cellcolor{gray!5}  & \cellcolor{gray!5}  & \cellcolor{gray!5}  & \cellcolor{gray!5}  & \cellcolor{gray!5} 0.43 & \cellcolor{gray!5}  \\ 
  \cellcolor{gray!5} edm & \cellcolor{gray!5} STA & \cellcolor{gray!5} $\mathbf{0.38}$ & \cellcolor{gray!5} $\mathbf{0.44}$ & \cellcolor{gray!5}  & \cellcolor{gray!5}  & \cellcolor{gray!5}  & \cellcolor{gray!5}  & \cellcolor{gray!5}  & \cellcolor{gray!5} $\mathbf{0.41}$ & \cellcolor{gray!5}  \\ 
  \cellcolor{gray!5} edm & \cellcolor{gray!5} CMOB & \cellcolor{gray!5} 0.4 & \cellcolor{gray!5} 0.45 & \cellcolor{gray!5}  & \cellcolor{gray!5}  & \cellcolor{gray!5}  & \cellcolor{gray!5}  & \cellcolor{gray!5}  & \cellcolor{gray!5} 0.43 & \cellcolor{gray!5}  \\ 
  \cellcolor{gray!25} enb & \cellcolor{gray!25} IM & \cellcolor{gray!25} 0.01 & \cellcolor{gray!25} $\mathbf{0.04}$ & \cellcolor{gray!25}  & \cellcolor{gray!25}  & \cellcolor{gray!25}  & \cellcolor{gray!25}  & \cellcolor{gray!25}  & \cellcolor{gray!25} $\mathbf{0.02}$ & \cellcolor{gray!25}  \\ 
  \cellcolor{gray!25} enb & \cellcolor{gray!25} STA & \cellcolor{gray!25} $\mathbf{0}$ & \cellcolor{gray!25} $\mathbf{0.04}$ & \cellcolor{gray!25}  & \cellcolor{gray!25}  & \cellcolor{gray!25}  & \cellcolor{gray!25}  & \cellcolor{gray!25}  & \cellcolor{gray!25} $\mathbf{0.02}$ & \cellcolor{gray!25}  \\ 
  \cellcolor{gray!25} enb & \cellcolor{gray!25} CMOB & \cellcolor{gray!25} 0.01 & \cellcolor{gray!25} $\mathbf{0.04}$ & \cellcolor{gray!25}  & \cellcolor{gray!25}  & \cellcolor{gray!25}  & \cellcolor{gray!25}  & \cellcolor{gray!25}  & \cellcolor{gray!25} $\mathbf{0.02}$ & \cellcolor{gray!25}  \\ 
  \cellcolor{gray!5} jura & \cellcolor{gray!5} IM & \cellcolor{gray!5} 0.45 & \cellcolor{gray!5} 0.27 & \cellcolor{gray!5} 0.38 & \cellcolor{gray!5}  & \cellcolor{gray!5}  & \cellcolor{gray!5}  & \cellcolor{gray!5}  & \cellcolor{gray!5} 0.37 & \cellcolor{gray!5}  \\ 
  \cellcolor{gray!5} jura & \cellcolor{gray!5} STA & \cellcolor{gray!5} 0.46 & \cellcolor{gray!5} $\mathbf{0.25}$ & \cellcolor{gray!5} $\mathbf{0.34}$ & \cellcolor{gray!5}  & \cellcolor{gray!5}  & \cellcolor{gray!5}  & \cellcolor{gray!5}  & \cellcolor{gray!5} $\mathbf{0.35}$ & \cellcolor{gray!5}  \\ 
  \cellcolor{gray!5} jura & \cellcolor{gray!5} CMOB & \cellcolor{gray!5} $\mathbf{0.44}$ & \cellcolor{gray!5} 0.26 & \cellcolor{gray!5} 0.35 & \cellcolor{gray!5}  & \cellcolor{gray!5}  & \cellcolor{gray!5}  & \cellcolor{gray!5}  & \cellcolor{gray!5} $\mathbf{0.35}$ & \cellcolor{gray!5}  \\ 
  \cellcolor{gray!25} scpf & \cellcolor{gray!25} IM & \cellcolor{gray!25} $\mathbf{1.29}$ & \cellcolor{gray!25} $\mathbf{0.6}$ & \cellcolor{gray!25} $\mathbf{1.53}$ & \cellcolor{gray!25}  & \cellcolor{gray!25}  & \cellcolor{gray!25}  & \cellcolor{gray!25}  & \cellcolor{gray!25} $\mathbf{1.14}$ & \cellcolor{gray!25}  \\ 
  \cellcolor{gray!25} scpf & \cellcolor{gray!25} STA & \cellcolor{gray!25} 1.31 & \cellcolor{gray!25} 0.62 & \cellcolor{gray!25} 1.6 & \cellcolor{gray!25}  & \cellcolor{gray!25}  & \cellcolor{gray!25}  & \cellcolor{gray!25}  & \cellcolor{gray!25} 1.18 & \cellcolor{gray!25}  \\ 
  \cellcolor{gray!25} scpf & \cellcolor{gray!25} CMOB & \cellcolor{gray!25} $\mathbf{1.29}$ & \cellcolor{gray!25} 0.61 & \cellcolor{gray!25} 1.58 & \cellcolor{gray!25}  & \cellcolor{gray!25}  & \cellcolor{gray!25}  & \cellcolor{gray!25}  & \cellcolor{gray!25} 1.16 & \cellcolor{gray!25}  \\ 
  \cellcolor{gray!5} sf1 & \cellcolor{gray!5} IM & \cellcolor{gray!5} 1.05 & \cellcolor{gray!5} 1.16 & \cellcolor{gray!5} 1.29 & \cellcolor{gray!5}  & \cellcolor{gray!5}  & \cellcolor{gray!5}  & \cellcolor{gray!5}  & \cellcolor{gray!5} 1.17 & \cellcolor{gray!5}  \\ 
  \cellcolor{gray!5} sf1 & \cellcolor{gray!5} STA & \cellcolor{gray!5} 1.02 & \cellcolor{gray!5} 1.08 & \cellcolor{gray!5} 1.18 & \cellcolor{gray!5}  & \cellcolor{gray!5}  & \cellcolor{gray!5}  & \cellcolor{gray!5}  & \cellcolor{gray!5} 1.09 & \cellcolor{gray!5}  \\ 
  \cellcolor{gray!5} sf1 & \cellcolor{gray!5} CMOB & \cellcolor{gray!5} $\mathbf{1.01}$ & \cellcolor{gray!5} $\mathbf{1.07}$ & \cellcolor{gray!5} $\mathbf{1.17}$ & \cellcolor{gray!5}  & \cellcolor{gray!5}  & \cellcolor{gray!5}  & \cellcolor{gray!5}  & \cellcolor{gray!5} $\mathbf{1.08}$ & \cellcolor{gray!5}  \\ 
  \cellcolor{gray!25} sf2 & \cellcolor{gray!25} IM & \cellcolor{gray!25} 0.98 & \cellcolor{gray!25} 1.12 & \cellcolor{gray!25} 1.34 & \cellcolor{gray!25}  & \cellcolor{gray!25}  & \cellcolor{gray!25}  & \cellcolor{gray!25}  & \cellcolor{gray!25} 1.14 & \cellcolor{gray!25}  \\ 
  \cellcolor{gray!25} sf2 & \cellcolor{gray!25} STA & \cellcolor{gray!25} $\mathbf{0.93}$ & \cellcolor{gray!25} 1.1 & \cellcolor{gray!25} $\mathbf{1.32}$ & \cellcolor{gray!25}  & \cellcolor{gray!25}  & \cellcolor{gray!25}  & \cellcolor{gray!25}  & \cellcolor{gray!25} $\mathbf{1.12}$ & \cellcolor{gray!25}  \\ 
  \cellcolor{gray!25} sf2 & \cellcolor{gray!25} CMOB & \cellcolor{gray!25} 0.98 & \cellcolor{gray!25} $\mathbf{1.07}$ & \cellcolor{gray!25} 1.39 & \cellcolor{gray!25}  & \cellcolor{gray!25}  & \cellcolor{gray!25}  & \cellcolor{gray!25}  & \cellcolor{gray!25} 1.15 & \cellcolor{gray!25}  \\ 
  \cellcolor{gray!5} slump & \cellcolor{gray!5} IM & \cellcolor{gray!5} $\mathbf{0.7}$ & \cellcolor{gray!5} $\mathbf{0.62}$ & \cellcolor{gray!5} 0.31 & \cellcolor{gray!5}  & \cellcolor{gray!5}  & \cellcolor{gray!5}  & \cellcolor{gray!5}  & \cellcolor{gray!5} 0.55 & \cellcolor{gray!5}  \\ 
  \cellcolor{gray!5} slump & \cellcolor{gray!5} STA & \cellcolor{gray!5} 0.76 & \cellcolor{gray!5} 0.64 & \cellcolor{gray!5} 0.26 & \cellcolor{gray!5}  & \cellcolor{gray!5}  & \cellcolor{gray!5}  & \cellcolor{gray!5}  & \cellcolor{gray!5} 0.55 & \cellcolor{gray!5}  \\ 
  \cellcolor{gray!5} slump & \cellcolor{gray!5} CMOB & \cellcolor{gray!5} 0.75 & \cellcolor{gray!5} 0.63 & \cellcolor{gray!5} $\mathbf{0.22}$ & \cellcolor{gray!5}  & \cellcolor{gray!5}  & \cellcolor{gray!5}  & \cellcolor{gray!5}  & \cellcolor{gray!5} $\mathbf{0.53}$ & \cellcolor{gray!5}  \\ 
  \cellcolor{gray!25} emotions & \cellcolor{gray!25} IM & \cellcolor{gray!25} \textit{0.21} & \cellcolor{gray!25} \textit{\textbf{0.22}} & \cellcolor{gray!25} \textit{0.21} & \cellcolor{gray!25} \textit{\textbf{0.09}} & \cellcolor{gray!25} \textit{\textbf{0.17}} & \cellcolor{gray!25} \textit{\textbf{0.16}} & \cellcolor{gray!25}  & \cellcolor{gray!25}  & \cellcolor{gray!25} \textit{\textbf{0.18}} \\ 
  \cellcolor{gray!25} emotions & \cellcolor{gray!25} STA & \cellcolor{gray!25} \textit{\textbf{0.2}} & \cellcolor{gray!25} \textit{\textbf{0.22}} & \cellcolor{gray!25} \textit{\textbf{0.2}} & \cellcolor{gray!25} \textit{\textbf{0.09}} & \cellcolor{gray!25} \textit{\textbf{0.17}} & \cellcolor{gray!25} \textit{0.17} & \cellcolor{gray!25}  & \cellcolor{gray!25}  & \cellcolor{gray!25} \textit{\textbf{0.18}} \\ 
  \cellcolor{gray!25} emotions & \cellcolor{gray!25} CMOB & \cellcolor{gray!25} \textit{\textbf{0.2}} & \cellcolor{gray!25} \textit{0.23} & \cellcolor{gray!25} \textit{0.22} & \cellcolor{gray!25} \textit{0.11} & \cellcolor{gray!25} \textit{\textbf{0.17}} & \cellcolor{gray!25} \textit{0.17} & \cellcolor{gray!25}  & \cellcolor{gray!25}  & \cellcolor{gray!25} \textit{\textbf{0.18}} \\ 
  \cellcolor{gray!5} image & \cellcolor{gray!5} IM & \cellcolor{gray!5} \textit{\textbf{0.13}} & \cellcolor{gray!5} \textit{\textbf{0.18}} & \cellcolor{gray!5} \textit{0.26} & \cellcolor{gray!5} \textit{\textbf{0.12}} & \cellcolor{gray!5} \textit{0.2} & \cellcolor{gray!5}  & \cellcolor{gray!5}  & \cellcolor{gray!5}  & \cellcolor{gray!5} \textit{0.18} \\ 
  \cellcolor{gray!5} image & \cellcolor{gray!5} STA & \cellcolor{gray!5} \textit{\textbf{0.13}} & \cellcolor{gray!5} \textit{\textbf{0.18}} & \cellcolor{gray!5} \textit{\textbf{0.25}} & \cellcolor{gray!5} \textit{\textbf{0.12}} & \cellcolor{gray!5} \textit{\textbf{0.19}} & \cellcolor{gray!5}  & \cellcolor{gray!5}  & \cellcolor{gray!5}  & \cellcolor{gray!5} \textit{\textbf{0.17}} \\ 
  \cellcolor{gray!5} image & \cellcolor{gray!5} CMOB & \cellcolor{gray!5} \textit{\textbf{0.13}} & \cellcolor{gray!5} \textit{0.21} & \cellcolor{gray!5} \textit{0.29} & \cellcolor{gray!5} \textit{\textbf{0.12}} & \cellcolor{gray!5} \textit{\textbf{0.19}} & \cellcolor{gray!5}  & \cellcolor{gray!5}  & \cellcolor{gray!5}  & \cellcolor{gray!5} \textit{0.19} \\ 
  \cellcolor{gray!25} reuters & \cellcolor{gray!25} IM & \cellcolor{gray!25} \textit{\textbf{0.07}} & \cellcolor{gray!25} \textit{\textbf{0.09}} & \cellcolor{gray!25} \textit{0.08} & \cellcolor{gray!25} \textit{\textbf{0.07}} & \cellcolor{gray!25} \textit{0.06} & \cellcolor{gray!25} \textit{\textbf{0.06}} & \cellcolor{gray!25} \textit{\textbf{0.06}} & \cellcolor{gray!25}  & \cellcolor{gray!25} \textit{\textbf{0.07}} \\ 
  \cellcolor{gray!25} reuters & \cellcolor{gray!25} STA & \cellcolor{gray!25} \textit{\textbf{0.07}} & \cellcolor{gray!25} \textit{\textbf{0.09}} & \cellcolor{gray!25} \textit{0.08} & \cellcolor{gray!25} \textit{\textbf{0.07}} & \cellcolor{gray!25} \textit{\textbf{0.05}} & \cellcolor{gray!25} \textit{\textbf{0.06}} & \cellcolor{gray!25} \textit{\textbf{0.06}} & \cellcolor{gray!25}  & \cellcolor{gray!25} \textit{\textbf{0.07}} \\ 
  \cellcolor{gray!25} reuters & \cellcolor{gray!25} CMOB & \cellcolor{gray!25} \textit{\textbf{0.07}} & \cellcolor{gray!25} \textit{\textbf{0.09}} & \cellcolor{gray!25} \textit{\textbf{0.07}} & \cellcolor{gray!25} \textit{\textbf{0.07}} & \cellcolor{gray!25} \textit{0.06} & \cellcolor{gray!25} \textit{0.08} & \cellcolor{gray!25} \textit{\textbf{0.06}} & \cellcolor{gray!25}  & \cellcolor{gray!25} \textit{\textbf{0.07}} \\ 
  \cellcolor{gray!5} scene & \cellcolor{gray!5} IM & \cellcolor{gray!5} \textit{\textbf{0.09}} & \cellcolor{gray!5} \textit{\textbf{0.03}} & \cellcolor{gray!5} \textit{\textbf{0.05}} & \cellcolor{gray!5} \textit{\textbf{0.05}} & \cellcolor{gray!5} \textit{\textbf{0.15}} & \cellcolor{gray!5} \textit{\textbf{0.12}} & \cellcolor{gray!5}  & \cellcolor{gray!5}  & \cellcolor{gray!5} \textit{\textbf{0.08}} \\ 
  \cellcolor{gray!5} scene & \cellcolor{gray!5} STA & \cellcolor{gray!5} \textit{0.1} & \cellcolor{gray!5} \textit{\textbf{0.03}} & \cellcolor{gray!5} \textit{\textbf{0.05}} & \cellcolor{gray!5} \textit{\textbf{0.05}} & \cellcolor{gray!5} \textit{\textbf{0.15}} & \cellcolor{gray!5} \textit{\textbf{0.12}} & \cellcolor{gray!5}  & \cellcolor{gray!5}  & \cellcolor{gray!5} \textit{\textbf{0.08}} \\ 
  \cellcolor{gray!5} scene & \cellcolor{gray!5} CMOB & \cellcolor{gray!5} \textit{\textbf{0.09}} & \cellcolor{gray!5} \textit{\textbf{0.03}} & \cellcolor{gray!5} \textit{\textbf{0.05}} & \cellcolor{gray!5} \textit{\textbf{0.05}} & \cellcolor{gray!5} \textit{\textbf{0.15}} & \cellcolor{gray!5} \textit{\textbf{0.12}} & \cellcolor{gray!5}  & \cellcolor{gray!5}  & \cellcolor{gray!5} \textit{\textbf{0.08}} \\ 
  \cellcolor{gray!25} youtube & \cellcolor{gray!25} IM & \cellcolor{gray!25} \textit{\textbf{0.13}} & \cellcolor{gray!25} 0.73 & \cellcolor{gray!25} 1.02 & \cellcolor{gray!25} $\mathbf{0.99}$ & \cellcolor{gray!25} 1.05 & \cellcolor{gray!25} $\mathbf{0.95}$ & \cellcolor{gray!25}  & \cellcolor{gray!25}  & \cellcolor{gray!25}  \\ 
  \cellcolor{gray!25} youtube & \cellcolor{gray!25} STA & \cellcolor{gray!25} \textit{\textbf{0.13}} & \cellcolor{gray!25} 0.73 & \cellcolor{gray!25} 1.02 & \cellcolor{gray!25} $\mathbf{0.99}$ & \cellcolor{gray!25} 1.06 & \cellcolor{gray!25} $\mathbf{0.95}$ & \cellcolor{gray!25}  & \cellcolor{gray!25}  & \cellcolor{gray!25}  \\ 
  \cellcolor{gray!25} youtube & \cellcolor{gray!25} CMOB & \cellcolor{gray!25} \textit{0.14} & \cellcolor{gray!25} $\mathbf{0.72}$ & \cellcolor{gray!25} $\mathbf{1}$ & \cellcolor{gray!25} $\mathbf{0.99}$ & \cellcolor{gray!25} $\mathbf{1.02}$ & \cellcolor{gray!25} $\mathbf{0.95}$ & \cellcolor{gray!25}  & \cellcolor{gray!25}  & \cellcolor{gray!25}  \\ 
  \cellcolor{gray!5} sens & \cellcolor{gray!5} IM & \cellcolor{gray!5} $\mathbf{0.91}$ & \cellcolor{gray!5} \textit{\textbf{0.28}} & \cellcolor{gray!5} 0.73 & \cellcolor{gray!5} 1.02 & \cellcolor{gray!5}  & \cellcolor{gray!5}  & \cellcolor{gray!5}  & \cellcolor{gray!5}  & \cellcolor{gray!5}  \\ 
  \cellcolor{gray!5} sens & \cellcolor{gray!5} STA & \cellcolor{gray!5} $\mathbf{0.91}$ & \cellcolor{gray!5} \textit{\textbf{0.28}} & \cellcolor{gray!5} 0.73 & \cellcolor{gray!5} 1.02 & \cellcolor{gray!5}  & \cellcolor{gray!5}  & \cellcolor{gray!5}  & \cellcolor{gray!5}  & \cellcolor{gray!5}  \\ 
  \cellcolor{gray!5} sens & \cellcolor{gray!5} CMOB & \cellcolor{gray!5} 0.93 & \cellcolor{gray!5} \textit{0.3} & \cellcolor{gray!5} $\mathbf{0.72}$ & \cellcolor{gray!5} $\mathbf{1}$ & \cellcolor{gray!5}  & \cellcolor{gray!5}  & \cellcolor{gray!5}  & \cellcolor{gray!5}  & \cellcolor{gray!5}  \\ 
   \bottomrule
\end{tabular}
\caption{Benchmark results. Mean squared error (MSE) of standardized target variables for regression tasks and mean misclassification error (mmce) for classification tasks (in italic). The best performing algorithm per dataset and target is highlighted in bold font.} 
\label{tab.mtr}
\end{table}

\subsection{Interpretation of Target Dependencies}
\subsection*{Example: Andromeda Dataset}
The Andromeda dataset (andro) \cite{hatzikos2008empirical} deals with the prediction of water quality variables (\textit{temperature, pH, conductivity, salinity, oxygen, turbidity}).
CMOB performed well on this dataset and  made improvements for every target variable and performed almost as well as the stacking algorithm:
\begin{center}
\begin{tabular}{llrrrrrrr}
  \toprule
Dataset & Algorithm & temp. & pH & conductivity & salinity & oxygen & turbidity & MMSE \\ 
  \midrule
andro & IM & 0.12 & 0.26 & 0.18 & 0.18 & 0.39 & 0.4 & 0.26 \\ 
  andro & STA & $\mathbf{0.08}$ & $\mathbf{0.21}$ & $\mathbf{0.11}$ & $\mathbf{0.11}$ & $\mathbf{0.32}$ & $\mathbf{0.33}$ & $\mathbf{0.19}$ \\ 
  andro & CMOB & $\mathbf{0.08}$ & $\mathbf{0.21}$ & $\mathbf{0.11}$ & $\mathbf{0.11}$ & 0.35 & 0.36 & 0.2 \\ 
   \bottomrule
\end{tabular}

\end{center}

To further inspect the target dependencies, we plot the base learner coefficients for each target (for effect size and direction) together with the corresponding relative risk reduction  of the underlying boosting algorithm (for feature importance) in figure \ref{fig:figandro}.
The relative risk reduction $\text{vip}^{\text{rel}}_p$ of a base learner $p$ is the proportion of the base learner's risk reduction to the total risk reduction:
$$
\text{vip}^{\text{rel}}_p = \frac{\text{vip}_p}{\sum_{i = 1}^{\tau} \text{vip}_i}
$$
The numbers in the plots are the base learner's coefficients and the background color displays the relative risk reduction.

Figure \ref{fig:figandro} needs to be read row wise, e.g. for the target variable \textit{salinity}, the predicted targets \textit{conductivity} and \textit{salinity} have been selected by the boosting algorithm and both have a positive effect on the value of salinity. It is quite clear, that the predicted target value of the target itself should normally be the most important feature and should have a coefficient of around 1. However, we can see some anomalies, e.g. for the target \textit{turbidity}, the predicted target value of \textit{oxygen}, seems to be more important. A possible reason could be that the first prediction of the target \textit{turbidity} was not accurate in the first place. Nevertheless, we can also see that the resulting boosting models are quite sparse, since only a few base-learners were chosen for most target variables.


\begin{knitrout}
\definecolor{shadecolor}{rgb}{0.969, 0.969, 0.969}\color{fgcolor}\begin{figure}

{\centering \includegraphics[width=0.7\linewidth]{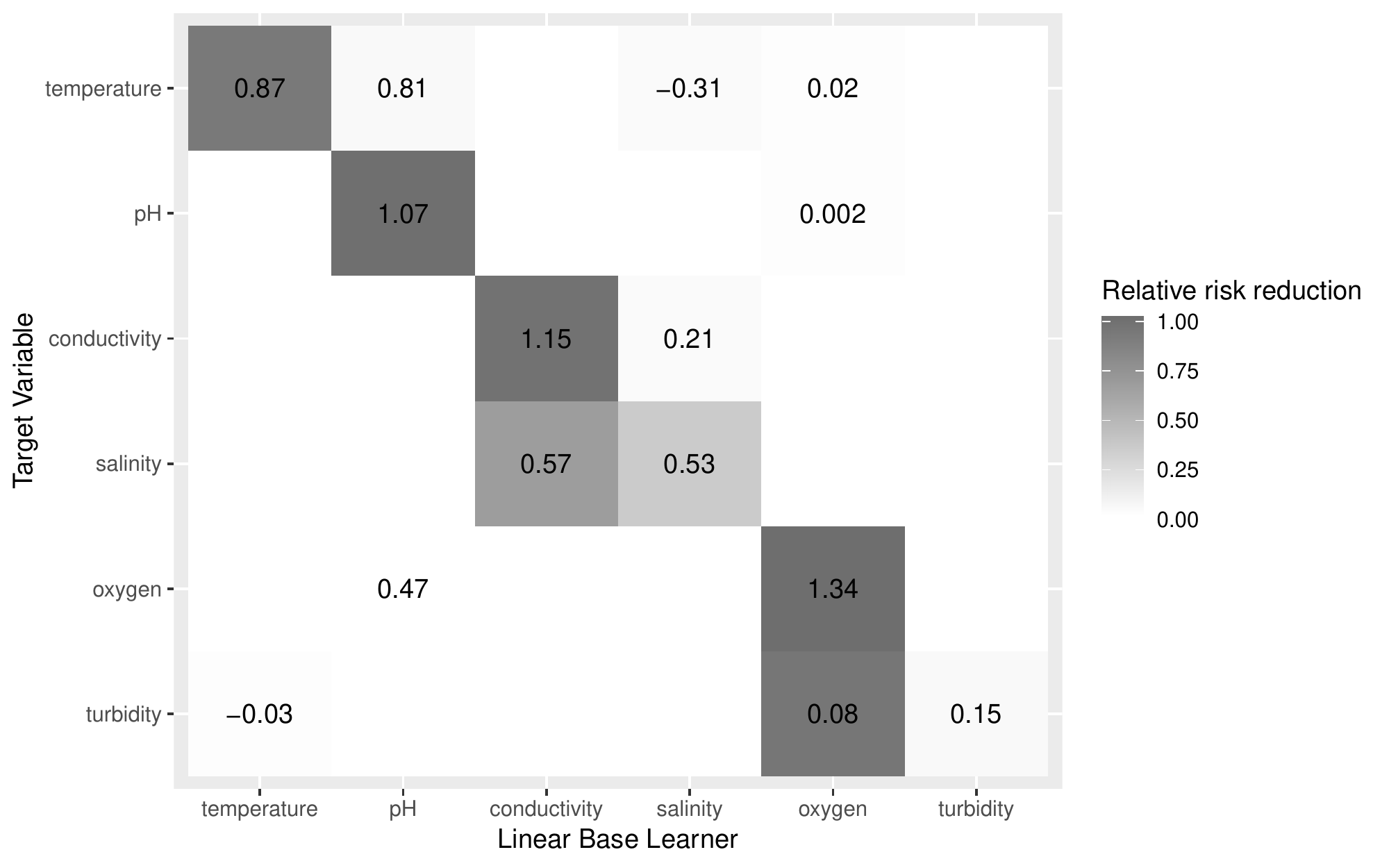} 

}

\caption{Coefficients of selected base-learners for each target variable of the Andromeda dataset. Example: The predicted target \textit{conductivity} has a positive effect ($\theta = 0.57$) on the target variable \textit{salinity} and also had a relatively high feature importance because of the high relative risk reduction.}\label{fig:figandro}
\end{figure}

\end{knitrout}

\subsection*{Example: Slump Dataset}
The Slump \cite{yeh2007modeling} dataset deals with the prediction of three properties of concrete (\textit{slump}, \textit{flow} and \textit{compressive strength}). CMOB and STA could make considerable improvements in the prediction of the target \textit{compressive strength}:
\begin{center}
\begin{tabular}{llrrcr}
  \toprule
Dataset & Algorithm & slump & flow & compressive strength & MMSE \\ 
  \midrule
slump & IM & $\mathbf{0.7}$ & $\mathbf{0.62}$ & 0.31 & 0.55 \\ 
  slump & STA & 0.76 & 0.64 & 0.26 & 0.55 \\ 
  slump & CMOB & 0.75 & 0.63 & $\mathbf{0.22}$ & $\mathbf{0.53}$ \\ 
   \bottomrule
\end{tabular}

\end{center}

One might argue that the improvements are due to exploiting target dependencies.
But if we look more closely at the selected base-learners of CMOB (see figure \ref{fig:slump}) we can see that the only base-learner chosen for the target \textit{compressive strength} is the target itself. This was also often the case for the models in the cross-validation iterations. Other targets were rarely chosen and had small coefficients.
Interestingly, we could achieve a performance improvement only by linearly transforming the predictions of the target \textit{compressive strength}.

\begin{knitrout}
\definecolor{shadecolor}{rgb}{0.969, 0.969, 0.969}\color{fgcolor}\begin{figure}

{\centering \includegraphics[width=0.7\linewidth]{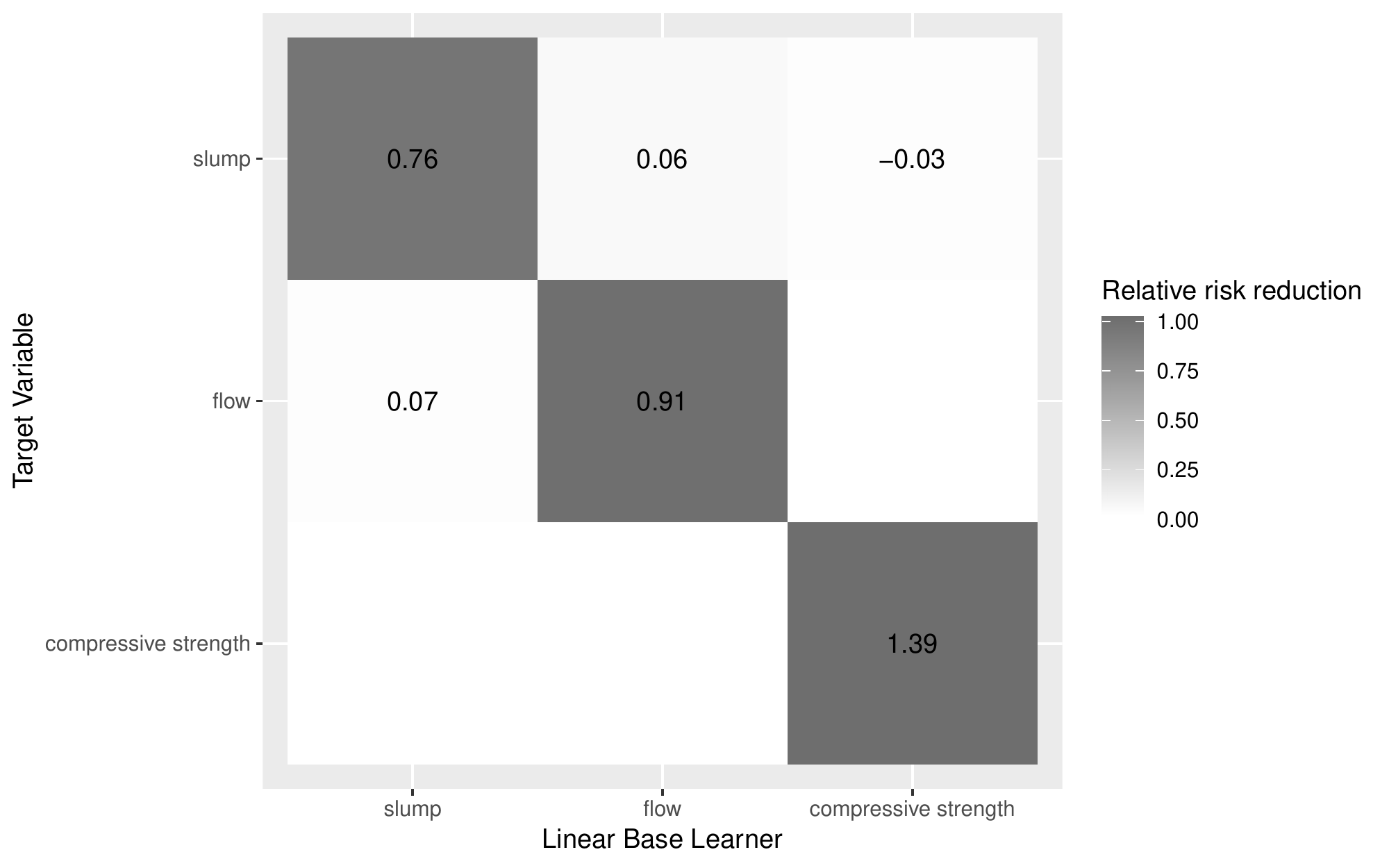} 

}

\caption[Coefficients of selected base-learners for each target variable of the Slump dataset]{Coefficients of selected base-learners for each target variable of the Slump dataset. The feature importance of the base-learners is highlighted by the background color.}\label{fig:slump}
\end{figure}

\end{knitrout}

\section{Conclusion and Outlook}
\label{sec:summary}

In this paper we defined the problem transformation method for multi-output prediction problems of possibly mixed target spaces.
We introduced a novel algorithm CMOB (component-wise multi-output boosting) which simultaneously learns dependencies within target variables in a sparse and interpretable manner.
Through a benchmark experiment with real-world datasets, we showed that, at least for some datasets, the performance of CMOB was comparable to the stacking method's performance (STA). In contrast to STA, which trains (possibly black-box) machine-learning models in a second step, CMOB learns the target dependencies with an inherently interpretable model.
With the help of CMOB, we were able to find an example, where improvements of predictive performance could be made for one target without using information of other targets. This would otherwise have been attributed to the exploitation of target dependencies.

We limited the choice of datasets to a rather small number of targets (less than 7).
Future work should address investigations of the performance of CMOB on datasets with many targets.
Since CMOB tries to model target dependencies in a sparse manner, this could be an advantage over STA, which, depending on the choice of the underlying machine learning models, cannot handle noisy variables very well.

\subsection*{Acknowledgements}
This work has been partially supported by the German Federal Ministry of Education and Research (BMBF) under Grant No. 01IS18036A. The authors of this work take full responsibilities for its content.
\clearpage
\bibliographystyle{splncs04}
\bibliography{Bib}

\begin{thebibliography}{10}
\providecommand{\url}[1]{\texttt{#1}}
\providecommand{\urlprefix}{URL }
\providecommand{\doi}[1]{https://doi.org/#1}

\bibitem{Au2019}
Au, Q.: {Benchmark Code for: Component-Wise Boosting of Targets for
  Multi-Output Prediction}  (4 2019). \doi{10.6084/m9.figshare.7957292.v2}

\bibitem{biel2013youtube}
Biel, J.I., Gatica-Perez, D.: The youtube lens: Crowdsourced personality
  impressions and audiovisual analysis of vlogs. Multimedia, IEEE Transactions
  on  \textbf{15}(1),  41--55 (2013)

\bibitem{Borchani2015}
Borchani, H., Varando, G., Bielza, C., Larra{\~{n}}aga, P.: {A survey on
  multi-output regression}. Wiley Interdisciplinary Reviews: Data Mining and
  Knowledge Discovery  \textbf{5}(5),  216--233 (2015)

\bibitem{Boutell2003}
Boutell, M., Shen, X., Luo, J., Brown, C.: {Multi-label semantic scene
  classification}. Pattern Recognition. v37 i9  \textbf{1771},  1--26 (2003)

\bibitem{breiman2001random}
Breiman, L.: Random forests. Machine learning  \textbf{45}(1),  5--32 (2001)

\bibitem{buhlmann2003boosting}
B{\"u}hlmann, P., Yu, B.: Boosting with the l 2 loss: regression and
  classification. Journal of the American Statistical Association
  \textbf{98}(462),  324--339 (2003)

\bibitem{OpenMLR2017}
Casalicchio, G., Bossek, J., Lang, M., Kirchhoff, D., Kerschke, P., Hofner, B.,
  Seibold, H., Vanschoren, J., Bischl, B.: Openml: An r package to connect to
  the machine learning platform openml. Computational Statistics
  \textbf{32}(3),  1--15 (2017)

\bibitem{Chapman2007}
Chapman, B.P., Duberstein, P.R., S{\"o}rensen, S., Lyness, J.M.: Gender
  differences in five factor model personality traits in an elderly cohort.
  Personality and individual differences  \textbf{43}(6),  1594--1603 (2007)

\bibitem{Donnellan2008}
Donnellan, M.B., Lucas, R.E.: Age differences in the big five across the life
  span: evidence from two national samples. Psychology and aging
  \textbf{23}(3), ~558 (2008)

\bibitem{hatzikos2008empirical}
Hatzikos, E.V., Tsoumakas, G., Tzanis, G., Bassiliades, N., Vlahavas, I.: An
  empirical study on sea water quality prediction. Knowledge-Based Systems
  \textbf{21}(6),  471--478 (2008)

\bibitem{rfsrc2019}
Ishwaran, H., Kogalur, U.: Random Forests for Survival, Regression, and
  Classification (RF-SRC) (2019),
  \url{https://cran.r-project.org/package=randomForestSRC}, r package version
  2.8.0

\bibitem{Kosinski2013}
Kosinski, M., Stillwell, D., Graepel, T.: Private traits and attributes are
  predictable from digital records of human behavior. Proceedings of the
  National Academy of Sciences  \textbf{110}(15),  5802--5805 (2013)

\bibitem{Li2012}
Li, T.: {Detecting Emotion in Text} (November 2003) (2012)

\bibitem{LozaMencia2016}
{Loza Menc{\'{i}}a}, E., Janssen, F.: {Learning rules for multi-label
  classification: a stacking and a separate-and-conquer approach}. Machine
  Learning  \textbf{105}(1),  77--126 (oct 2016)

\bibitem{molnar2019}
Molnar, C.: Interpretable Machine Learning (2019)

\bibitem{Montanes2014}
Monta{\~{n}}es, E., Senge, R., Barranquero, J., {Ram{\'{o}}n Quevedo}, J.,
  {Jos{\'{e}} Del Coz}, J., H{\"{u}}llermeier, E.: {Dependent binary relevance
  models for multi-label classification}. Pattern Recognition  \textbf{47}(3),
  1494--1508 (2014)

\bibitem{Probst2017}
Probst, P., Au, Q., Casalicchio, G., Stachl, C., Bischl, B.: {Multilabel
  Classification with R Package mlr}. {The R Journal}  \textbf{9}(1),  352--369
  (2017)

\bibitem{Read2011}
Read, J., Pfahringer, B., Holmes, G., Frank, E., {Brodley Read}, C.J.,
  Pfahringer, B., Holmes, G., Frank, E., Read, J.: {Classifier chains for
  multi-label classification}. Mach Learn  \textbf{85},  333--359 (2011)

\bibitem{Read2016}
Read, J., Reutemann, P., Pfahringer, B., Holmes, G.: {MEKA: A
  Multi-label/Multi-target extension to WEKA}. Journal of Machine Learning
  Reasearch. Available at: http://jmlr.org/papers/volume17/12-164/12-164.pdf
  \textbf{17}(21), ~1--5 (2016)

\bibitem{Schapire2000}
Schapire, R.E., Singer, Y.: {BoosTexter: A Boosting-based System for Text
  Categorization}. Machine Learning  \textbf{39},  135--168 (2000)

\bibitem{schmid2008boosting}
Schmid, M., Hothorn, T.: Boosting additive models using component-wise
  p-splines. Computational Statistics \& Data Analysis  \textbf{53}(2),
  298--311 (2008)

\bibitem{Schoedel2018}
Schoedel, R., Au, Q., V{\"{o}}lkel, S.T., Lehmann, F., Becker, D.,
  B{\"{u}}hner, M., Bischl, B., Hussmann, H., Stachl, C.: {Digital Footprints
  of Sensation Seeking}. Zeitschrift f{\"{u}}r Psychologie  \textbf{226}(4),
  232--245 (2018)

\bibitem{Senge}
Senge, R., {Jos{\'{e}} Del Coz}, J., Hüllermeier, E.: {Rectifying Classifier
  Chains for Multi-Label Classification}. Tech. rep.

\bibitem{Shi2012}
Shi, C., Kong, X., Yu, P.S., Wang, B.: {Multi-Objective Multi-Label
  Classification}. Proceedings of the 2012 SIAM International Conference on
  Data Mining pp. 355--366 (2012)

\bibitem{SpyromitrosXioufis2012}
Spyromitros-Xioufis, E., Tsoumakas, G., Groves, W., Vlahavas, I.: {Multi-Target
  Regression via Input Space Expansion: Treating Targets as Inputs}. Machine
  Learning  \textbf{104}(1),  55--98 (nov 2012)

\bibitem{Spyromitros-Xioufis2016}
Spyromitros-Xioufis, E., Tsoumakas, G., Groves, W., Vlahavas, I.: {Multi-target
  regression via input space expansion: treating targets as inputs}. Machine
  Learning  \textbf{104}(1),  55--98 (2016)

\bibitem{Tsoumakas2007}
Tsoumakas, G., Katakis, I.: {Multi-label classification: An overview}.
  International Journal of Data Warehousing and Mining  \textbf{3},  1--13
  (2007)

\bibitem{mulan}
Tsoumakas, G., Spyromitros-Xioufis, E., Vilcek, J., Vlahavas, I.: Mulan: A java
  library for multi-label learning. Journal of Machine Learning Research
  \textbf{12},  2411--2414 (2011)

\bibitem{OpenML2013}
Vanschoren, J., van Rijn, J.N., Bischl, B., Torgo, L.: Openml: Networked
  science in machine learning. SIGKDD Explorations  \textbf{15}(2),  49--60
  (2013)

\bibitem{Waegeman2018}
Waegeman, W., Dembczynski, K., Huellermeier, E.: {Multi-Target Prediction: A
  Unifying View on Problems and Methods}  (sep 2018)

\bibitem{Xu2019ASO}
Xu, D., Shi, Y., Tsang, I.W., Ong, Y.S., Gong, C., Shen, X.: A survey on
  multi-output learning. CoRR  \textbf{abs/1901.00248} (2019)

\bibitem{yeh2007modeling}
Yeh, I.C.: Modeling slump flow of concrete using second-order regressions and
  artificial neural networks. Cement and Concrete Composites  \textbf{29}(6),
  474--480 (2007)

\bibitem{Zhang2007}
Zhang, M.L., Zhou, Z.H.: {ML-KNN: A lazy learning approach to multi-label
  learning}. Pattern Recognition  \textbf{40}(7),  2038--2048 (jul 2007)

\bibitem{Zhang2014}
Zhang, M.L., Zhou, Z.H.: {A review on multi-label learning algorithms} (2014)

\end{thebibliography}
\end{document}